\documentclass{article}

\usepackage{PRIMEarxiv}

\usepackage[utf8]{inputenc} 
\usepackage[T1]{fontenc}    
\usepackage{hyperref}       
\usepackage{url}            
\usepackage{booktabs}       
\usepackage{amsfonts}       
\usepackage{amsmath}
\usepackage{amssymb}
\usepackage{bm}
\usepackage{nicefrac}       
\usepackage{microtype}      
\usepackage{lipsum}
\usepackage{fancyhdr}       
\usepackage{graphicx}       
\graphicspath{{Figures/}}     

\usepackage{mathtools}
\usepackage{subcaption}
\newtheorem{definition}{Definition}
\DeclarePairedDelimiter\abs{\lvert}{\rvert}
\makeatletter
\let\oldabs\abs
\def\abs{\@ifstar{\oldabs}{\oldabs*}}
\makeatother

\title{Admissable: Training Reinforcement Learning Agents against Adversarial
Missingness}

\author{
  Paul Stahlhofen, Luca Hermes, Tim Kochs, Markus Vieth, Barbara Hammer \\
  Bielefeld University \\
\texttt{\{pstahlhofen,lhermes,tkochs,mvieth,bhammer\}@techfak.uni-bielefeld.de} \\
}

\begin{document}
\maketitle

\begin{abstract}
In order to make Reinforcement Learning algorithms applicable in real world
scenarios, safety must be ensured even under adverse operating conditions. In
this work, we consider the challenge of adversarial feature missingness: a
scenario in which an adversary occludes features from the agent's observation
in order to reduce performance as much as possible. We formally define
adversarial missingness for Reinforcement Learning and compare it to the
related concepts of $\ell_\infty$-norm bounded adversarial perturbations and
learning with missing data. We develop an adversarial
training algorithm and show its effectiveness in increasing robustness against
adversarial missingness on three MuJoCo benchmark environments. Compared to a
baseline trained with random uniform missingness, our method achieves better
robustness on all three tasks.
\end{abstract}


\section{Introduction}
Reinforcement Learning (RL) has shown great potential in many areas, from mastering
complex games \cite{silver_mastering_2017} to robotics \cite{Ha.etal_2020} and cooling of data centers \cite{luo_controlling_2022}. For deployment in
practise, the European Union defines strict guidelines on Trustworthy AI \cite{european_commission_directorate_general_for_communications_networks_content_and_technology_ethics_2019}. To meet the criteria for trustworthiness, RL agents must be trained for a robust
performance under adverse operating conditions. In this work, we consider a
scenario in which the most important part of the state observation is hidden
from the agent. We model this by introducing an
adversary with the goal of minimizing the agent's return by masking a limited
number of the state features. We refer to this scenario as \textit{Adversarial Missingness}
- a concept known from the field of causal structure learning  \cite{koyuncu_adversarial_2024}. Adversarial
Missingness bridges the gap between learning with missing data and learning
under adversarial attacks.

\textbf{RL with Missing Data:} Several approaches have been developed to
handle missing data in Reinforcement Learning. Most methods make use of an
imputation for the missing data, before training the agent on the imputed
observations \cite{fleming_missingness_2019} \cite{xiang_partially_2026}
\cite{mei_reinforcement_2023}. The algorithm proposed by
\cite{skand_simple_2024} groups features into modalities and uses a
transformer over the sequence of these modalities as part of the encoder. The
authors demonstrate the effectiveness of their approach even under multiple
missing modalities. \cite{wang_robust_2019} model the problem as a Partially
Observable Markov Decision Process (POMDP). They learn the environment's
transition dynamics under incomplete and noisy observations, employing a
model-based RL approach to optimize the policy.
\cite{becker_uncertainty_2022} develop an extension of the Recurrent State
Space Models (RSSMs) \cite{hafner_learning_2019} to better model aleatoric
uncertainty. They demonstrate that their world model can be used successfully
for the imputation of missing data. Although these works cover different
assumptions about data missingness, none of them discusses a scenario where the missing features are controlled by an adversary.

\textbf{RL with Adversarially Attacked States:}
Neural networks were shown to be vulnerable to adversarial attacks i.e. to
inputs that an attacker has intentionally created to cause the model to make a
mistake \cite{szegedy_intriguing_2014}. Though originally discovered in image
classification, the vulnerability of strong function approximators can have
catastrophic consequences also in Reinforcement Learning. \cite{huang_adversarial_2017} 
feed adversarial examples to the policies of image-based RL agents, showing
reduced performance for multiple learning algorithms. \cite{pattanaik_robust_2017} show
effectiveness of adversarial attacks also for smaller environments with state
representations that are not based on images. \cite{zhang_robust_2021-1}
introduce the State Adversarial Markov Decision Process (SA-MDP) to
theoretically analyze the problem and train for robustness against a learned
optimal adversary in a follow-up work \cite{zhang_robust_2021}. For a general
survey on robustness in Reinforcement Learning, we refer the interested reader
to \cite{moos_robust_2022}.\\
Most of these works consider adversarial inputs as perturbed original states
with a perturbation budget bounded in the $\ell_\infty$-norm. This is inspired by
image classification, where the classifier output should not change by making
imperceptible changes to every single pixel. However, as pointed out by
\cite{koyuncu_adversarial_2024}, these attacks cannot capture scenarios where
the adversary is only able to remove part of the observation, instead of
changing every single feature of the input. The same argument applies to other
$\ell_p$-norm bounded attacks with $p \geq 1$. In practise, it is debatable
whether the more likely threat model to a trained robot is one where a hacker
gets to manipulate all of its sensor values by a tiny amount, or one where the
attacker gets to remove or destroy a limited number of sensors, trying to pick
those on which the robot will likely rely the most. An important property of
standard adversarials in image recognition is being intuitively
indistinguishable for a human beholder. This property can admittedly get lost
in human-perceivable domains like vision or audio, if part of the input is
simply removed. On the other hand, many RL applications rely on features
measured by sensors that can hardly be monitored by humans as easily as
viewing an image. The notion of "imperceptibility" changes for these settings,
making missingness a potentially hidden threat as well.\\
Two related publications consider both missingness and adversarial attacks: 
\cite{li_robustlight_2025} employ a hierarchical Reinforcement Learning
approach for traffic signal control. They train one agent for the actual
control problem and a second one to produce input reconstructions via a
diffusion model to counter corruptions of the input induced by attacks or
missing data. However, they don't let the adversary determine which feature is
missing. \cite{kumar_policy_2022} develop a method to obtain provably robust
policies in a POMDP setting where the agent does not observe the full state.
They bound the perturbation of their adversary by the $\ell_2$-norm, hence not
allowing it to simply mask a feature from the agent's observation.

\section{Adversarial Missingness in Reinforcement Learning}
\label{sec:formalization}
As a basis, recall the standard definition of a Markov Decision Process
\begin{definition}{Markov Decision Process (MDP)}
A Markov Decision Process is a quintuplet $\mathcal{P} = (\mathcal{S},
\mathcal{A}, \mathcal{R}, \gamma, P_{\text{trans}})$ with a state space
$\mathcal{S} \subset \mathbb{R}^d$, an action space $\mathcal{A}$,
a set of possible rewards $\mathcal{R} \subseteq \mathbb{R}$, a discount
factor $\gamma \in [0, 1]$ and a transition function $P_{\text{trans}}(r,
s^\prime | s, a)$ mapping tuples of states and actions to probability
distributions over next states and rewards.
\end{definition}
At each time step, the agent observes a state $S_t$ and picks an
action $A_t$. Based on transition probability $P_{\text{trans}}$ the environment then
emits an immediate reward $R_{t+1}$ and transits to the next state $S_{t+1}$
until termination. The goal in standard Reinforcement Learning is to optimize
a policy $\pi(a|s)$ such that sampling actions from $\pi$ in each state
maximizes the expected cumulative reward

\begin{equation}
\max_{\pi} \mathbb{E}_{\pi, P_{\text{trans}}} \left[ \sum_{t=0}^T \gamma^t R_{t+1} \right]
\end{equation}

Equivalently, this objective can be phrased as maximizing the value of an
initial state, where the value function $v^\pi$ is defined as the expected
future reward for each state when following $\pi$ from that state onward
\begin{equation}
v^\pi(s) = \mathbb{E}_{P_{\text{trans}}, \pi} \left[ \sum_{k=0}^{T-(t+1)} \gamma^k R_{t+k+1} \big| S_t = s\right]\\
\end{equation}

We now extend the standard definition by adding an adversary that can remove
up to $k$ features from the state, for some $k<d$. Let $U:=\{A \subset \{1,
\hdots, d\} | \abs{A} \leq k\}$ contain all potentially missing subsets of
features. A parametrized missingness function $m$ can now be defined given $u \in U$

\begin{equation}
m_u(s)_i := \begin{cases}
\bot & \text{if}\ i \in u\\
s_i & \text{otherwise}
\end{cases}
\end{equation}

At each step, the agent now observes $m_u(S_t)$ instead of the full state
$S_t$ and has to take an action based on this limited information. An
important detail of our formalization is that $u$ is globally fixed, not
changing over state or time. That way, we can accurately model real-world
scenarios like sensor failures or sources of information that are permanently
removed. Analogous to standard MDPs, we can define an occluding value function

\begin{equation}
v_u^\pi(s) := \mathbb{E}_{P_{\text{trans}}, \pi, m_u} \left[ \sum_{k=0}^{T-(t+1)} \gamma^k R_{t+k+1} | S_t = s \right]\\
\end{equation}

The Bellman equation for the value function of this MDP with missingness is
given by
\begin{equation}
v_u^\pi(s) = \int_a \pi(a|m_u(s)) \int_r \int_{s^\prime} P_{\text{trans}}(r, s^\prime | s, a) [r + \gamma v_u^\pi(s^\prime)]
\end{equation}
The goal of the adversary is to minimize the value of the starting state by
picking a set of missing features
\begin{equation}
\min_{u \in U} v_u^\pi(s_0)
\label{eq:adversarial}
\end{equation}

Conversely, the objective of finding a robust policy against adversarial
missingness is expressed by
\begin{equation}
\max_\pi \min_{u \in U} v_u^\pi(s_0)
\label{eq:robust}
\end{equation}

\section{Adversarial Training against Missingness}

In this paper, we limit ourselves to experiments for the case where the
adversary may only remove one feature $k=1$. Further analysis for multiple
missing features is left for future work. Before trying to solve the
robustness objective, we have to find a suitable numerical representation for
partially occluded states. In order to introduce as little bias as possible,
we decide against a learned imputation and instead apply a masking
approach as used in \cite{skand_simple_2024}: all missing features are
replaced by zeros and a binary indicator vector $v$ with $v_i := \mathbb{I}(i
\in u)$ is concatenated to the state.

In their first paper discussing adversarial examples,
\cite{szegedy_intriguing_2014} propose adversarial training. They include
adversarial inputs in the training set to improve the robustness of their
classifier. In the following, we develop an adversarial training approach to
achieve robustness against adversarial missingness in RL (Eqn.
~\ref{eq:robust}). We use the Soft
Actor Critic (SAC) algorithm for training \cite{haarnoja_soft_2018}. SAC is a state of the
art off-policy algorithm, using an entropy maximization term in its loss for
improved exploration.
At each episode during training, we sample a missing feature from a probability distribution $p_{\text{miss}}$ over $U$.  
$p_{\text{miss}}$ is initialized as a uniform distribution and adapted to
simulate the adversary. Every $\zeta$ time steps, adversarial scores $a_u^\pi
:=\hat{v}_u^\pi(s_0)$ are determined for every $u \in U$, where $\hat{v}$ is
an approximation of the value function $v$ computed by running $N$ episodes and averaging the results. We update $p_{\text{miss}}$
according to the inverse adversarial scores
\begin{equation}
p_{\text{miss}}(i) = \frac{\frac{1}{a^\pi_u - l + \epsilon}}{\sum_{u \in U} \frac{1}{a^\pi_u - l + \epsilon}}
\end{equation}
where $l$ is a lower bound of the $a^\pi_u$ and $\epsilon$ is a small
tolerance to avoid division by zero. Using this update, features that lead to lower performance when
missing will be sampled more frequently by the adversary. We decide for a
stochastic adversary instead of a deterministic one that would always pick
$\min_u a^\pi_u$ in order to avoid training instabilities and forgetting. In cases where a lower bound is not known, we observe in practise
that good results can be achieved by estimating a bound and clipping values of
$a^\pi_u$ that fall below it. The values for $\zeta, \epsilon$ and $l$ used in
our experiments are given in Table ~\ref{tab:hyperparameters}. We compare our method to a baseline where $p_\text{miss}$ is fixed as a uniform distribution.

\begin{table}[ht]
\centering
\begin{tabular}{c|c|c|c}
Hyperparameter & Update Frequency & Tolerance & Return Lower Bound\\
\hline
Value & $\zeta=5000$ & $\epsilon=10^{-8}$ & $l=0$\\
\end{tabular}
\vspace{0.3cm}
\caption{Hyperparameters for Adversarial Training}
\label{tab:hyperparameters}
\end{table}

\section{Results and Discussion}
\begin{figure}[ht]
\centering
\begin{subfigure}{0.33\textwidth}
\includegraphics[width=\textwidth]{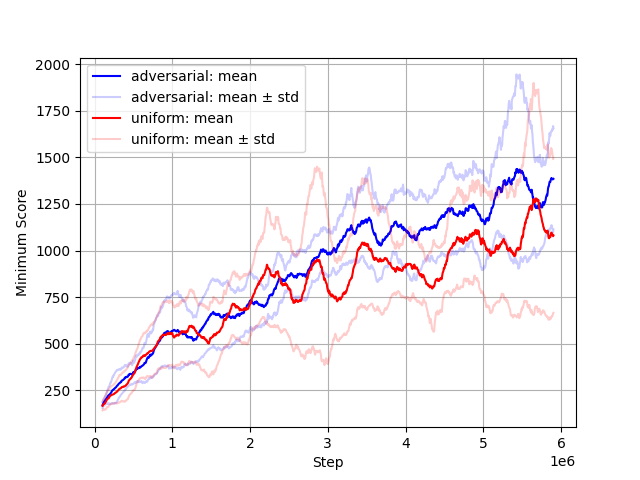}
\subcaption{Hopper-v5}
\end{subfigure}
\begin{subfigure}{0.33\textwidth}
\includegraphics[width=\textwidth]{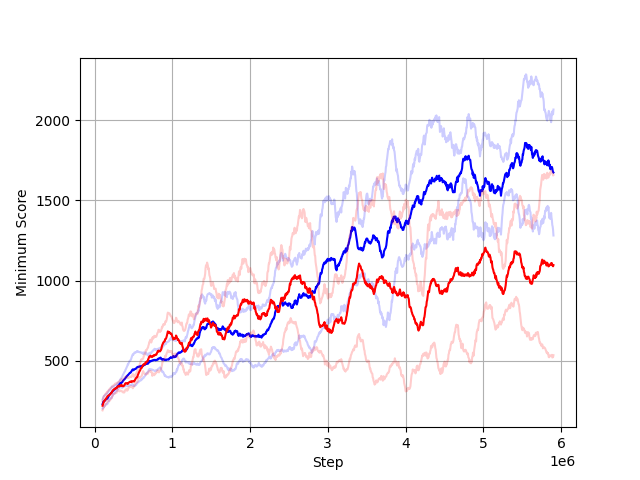}
\subcaption{Walker2d-v5}
\end{subfigure}
\begin{subfigure}{0.33\textwidth}
\includegraphics[width=\textwidth]{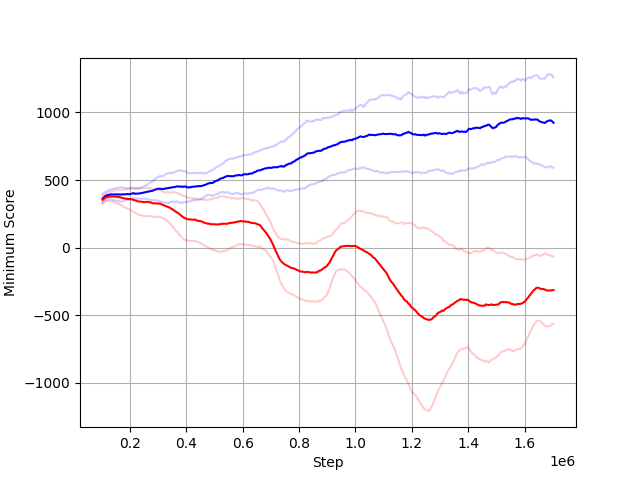}
\subcaption{Ant-v5}
\end{subfigure}
\caption{Performance under adversarial missingness. Every 5.000 time steps, we
determine the adversarial performance $\min_i a^\pi_i$ for training runs with
adversarial missingness (blue) and random uniform missingness (red). Results
are averaged over 5 runs and smoothed with a moving average over 40 records.}
\label{fig:adversarial_scores}
\end{figure}

We evaluate our adversarial training approach on three MuJoCo benchmark
environments: Hopper, Walker and Ant \cite{todorov_mujoco_2012}. The environments are structurally
similar. Given a robot with multiple joints, the task is to apply torque to
the joints in order to move forward as far as possible without falling. The
difference between the environments lies in the type of robot and the size of
the observation and action spaces (see Table ~\ref{tab:envs}). 
\begin{table}[ht]
\centering
\begin{tabular}{c|c|c|c}
 & Hopper & Walker & Ant\\
\hline
Number of Observations & 11 & 17 & 105\\
\hline
Number of Actions & 3 & 6 & 8\\
\end{tabular}
\vspace{0.3cm}
\caption{Properties of the MuJoCo environments used for training}
\label{tab:envs}
\end{table}
As we are interested in robustness against the strongest possible adversary,
we consider the minimum over the adversarial scores $a^\pi_u$ as a
performance measure. Intuitively speaking, this is the episodic
return\footnote{In our experiments, we compute the sum of rewards per episode
without discounting. This follows the standard implementation of policy
evaluation in Stable Baselines 3 \cite{raffin_stable-baselines3_2021}.} that an
agent can still achieve if the adversary managed to remove the most important
feature. Every time we update $p_{miss}$, we record the adversarial
performance. For the baseline with uniform missingness, we record adversarial
performance at the same intervals without updating the distribution. Results
are shown in Figure ~\ref{fig:adversarial_scores}.

We can see a better adversarial performance when training with adversarial
missingness in all three runs. The larger the observation space, the more
pronounced is the advantage. It appears that a more focused approach, forcing
the model to deal without the few features on which it may have overfitted so
far, becomes more important, the more features exist. In particular, note the
catastrophic adversarial performance of uniform missingness training in the
Ant environment. This makes a strong case for training with adversarial
missingness instead, as it shows that a model that has seemingly been
robustified against many possible missing features can still have a
tremendous weak spot. This kind of vulnerability might not be identified when
averaging performances across different missing features, if it is an outlier.
In order to enhance trustworthy AI and prevent catastrophic events caused by
sensor failures, we therefore recommend the use of adversarial missingness
instead of standard uniform missingness in training of Reinforcement Learning
agents.

\section{Conclusion}
In this work, we applied the concept of adversarial missingness to
Reinforcement Learning. We gave a formal definition of the problem, including
objectives for the adversary and for model robustness. After developing an
adversarial training approach to optimize the robustness objective, we
compared our method to a baseline with random uniform missingness on three
MuJoCo benchmark environments. We were able to show an improvement in
performance for all three cases, observing a higher benefit of adversarial
training for environments with larger observation spaces.
Adversarial missingness is an interesting and still mostly unexplored area,
leaving many directions for future work. Extending the scope to more than one
missing feature would lead to broader generalization. To that end, we plan to
explore more efficient ways of estimating adversarial scores and connections
to methods from Explainable AI.

\bibliographystyle{apalike}  
\bibliography{admissable}  

@misc{fleming_missingness_2019,
	title = {Missingness as {Stability}: {Understanding} the {Structure} of {Missingness} in {Longitudinal} {EHR} data and its {Impact} on {Reinforcement} {Learning} in {Healthcare}},
	shorttitle = {Missingness as {Stability}},
	url = {http://arxiv.org/abs/1911.07084},
	doi = {10.48550/arXiv.1911.07084},
	language = {en},
	urldate = {2026-05-11},
	publisher = {arXiv},
	author = {Fleming, Scott L. and Jeyapragasan, Kuhan and Duan, Tony and Ding, Daisy and Gombar, Saurabh and Shah, Nigam and Brunskill, Emma},
	month = nov,
	year = {2019},
	note = {arXiv:1911.07084 [cs]},
}

@misc{wang_robust_2019,
	title = {Robust {Reinforcement} {Learning} in {POMDPs} with {Incomplete} and {Noisy} {Observations}},
	url = {http://arxiv.org/abs/1902.05795},
	doi = {10.48550/arXiv.1902.05795},
	language = {en},
	urldate = {2026-05-11},
	publisher = {arXiv},
	author = {Wang, Yuhui and He, Hao and Tan, Xiaoyang},
	month = feb,
	year = {2019},
	note = {arXiv:1902.05795 [cs]},
}

@inproceedings{skand_simple_2024,
	title = {Simple {Masked} {Training} {Strategies} {Yield} {Control} {Policies} {That} {Are} {Robust} to {Sensor} {Failure}},
	url = {https://openreview.net/forum?id=AsbyZRdqPv},
	booktitle = {8th {Annual} {Conference} on {Robot} {Learning}},
	author = {Skand, Skand and Pandit, Bikram and Kim, Chanho and Fuxin, Li and Lee, Stefan},
	year = {2024},
}

@misc{kumar_policy_2022,
	title = {Policy {Smoothing} for {Provably} {Robust} {Reinforcement} {Learning}},
	url = {http://arxiv.org/abs/2106.11420},
	doi = {10.48550/arXiv.2106.11420},
	urldate = {2026-06-19},
	publisher = {arXiv},
	author = {Kumar, Aounon and Levine, Alexander and Feizi, Soheil},
	month = may,
	year = {2022},
	note = {arXiv:2106.11420 [cs.LG]},
}

@inproceedings{li_robustlight_2025,
	title = {{RobustLight}: {Improving} {Robustness} via {Diffusion} {Reinforcement} {Learning} for {Traffic} {Signal} {Control}},
	url = {https://openreview.net/forum?id=YGjd2xw98G},
	booktitle = {Forty-second {International} {Conference} on {Machine} {Learning}},
	author = {Li, Mingyuan and Wang, Jiahao and Yu, Guangsheng and Wang, Xu and Chen, Qianrun and Ni, Wei and Li, Lixiang and Peng, Haipeng},
	year = {2025},
}

@inproceedings{haarnoja_soft_2018,
	title = {Soft {Actor}-{Critic}: {Off}-{Policy} {Maximum} {Entropy} {Deep} {Reinforcement} {Learning} with a {Stochastic} {Actor}},
	volume = {80},
	shorttitle = {Soft {Actor}-{Critic}},
	url = {https://proceedings.mlr.press/v80/haarnoja18b},
	urldate = {2025-02-24},
	booktitle = {{PMLR}},
	author = {Haarnoja, Tuomas and Zhou, Aurick and Abbeel, Pieter and Levine, Sergey},
	month = aug,
	year = {2018},
	note = {arXiv:1801.01290 [cs]},
	pages = {1861--1870},
}

@article{koyuncu_adversarial_2024,
	title = {Adversarial {Missingness} {Attacks} on {Causal} {Structure} {Learning}},
	volume = {15},
	issn = {2157-6904, 2157-6912},
	url = {https://dl.acm.org/doi/10.1145/3682065},
	doi = {10.1145/3682065},
	language = {en},
	number = {6},
	urldate = {2026-03-19},
	journal = {ACM Transactions on Intelligent Systems and Technology},
	author = {Koyuncu, Deniz and Gittens, Alex and Yener, Bülent and Yung, Moti},
	month = dec,
	year = {2024},
	pages = {1--60},
}

@misc{szegedy_intriguing_2014,
	title = {Intriguing properties of neural networks},
	url = {http://arxiv.org/abs/1312.6199},
	doi = {10.48550/arXiv.1312.6199},
	urldate = {2026-09-02},
	publisher = {arXiv},
	author = {Szegedy, Christian and Zaremba, Wojciech and Sutskever, Ilya and Bruna, Joan and Erhan, Dumitru and Goodfellow, Ian and Fergus, Rob},
	month = feb,
	year = {2014},
	note = {arXiv:1312.6199 [cs.CV]},
}

@book{european_commission_directorate_general_for_communications_networks_content_and_technology_ethics_2019,
	address = {LU},
	title = {Ethics guidelines for trustworthy {AI}.},
	url = {https://data.europa.eu/doi/10.2759/346720},
	doi = {10.2759/346720},
	language = {eng},
	urldate = {2026-09-02},
	publisher = {Publications Office},
	author = {{European Commission}},
	year = {2019},
}

@article{xiang_partially_2026,
	title = {Partially observable reinforcement learning for blood glucose control under missing data},
	volume = {218},
	issn = {03608352},
	url = {https://linkinghub.elsevier.com/retrieve/pii/S0360835226003190},
	doi = {10.1016/j.cie.2026.112118},
	language = {en},
	urldate = {2026-09-02},
	journal = {Computers \& Industrial Engineering},
	author = {Xiang, Jiao and Kong, Nan and Yang, Ching-Chi and Luo, Li and Yu, Haiyan},
	month = aug,
	year = {2026},
	pages = {112118},
}

@inproceedings{mei_reinforcement_2023,
	title = {Reinforcement {Learning} {Approaches} for {Traffic} {Signal} {Control} under {Missing} {Data}},
	url = {http://arxiv.org/abs/2304.10722},
	doi = {10.24963/ijcai.2023/251},
	urldate = {2026-09-02},
	booktitle = {Proceedings of the {Thirty}-{Second} {International} {Joint} {Conference} on {Artificial} {Intelligence}},
	author = {Mei, Hao and Li, Junxian and Shi, Bin and Wei, Hua},
	month = aug,
	year = {2023},
	note = {arXiv:2304.10722 [cs.LG]},
	pages = {2261--2269},
}

@misc{zhang_robust_2021,
	title = {Robust {Reinforcement} {Learning} on {State} {Observations} with {Learned} {Optimal} {Adversary}},
	url = {http://arxiv.org/abs/2101.08452},
	urldate = {2024-06-04},
	author = {Zhang, Huan and Chen, Hongge and Boning, Duane and Hsieh, Cho-Jui},
	month = jan,
	year = {2021},
	note = {arXiv:2101.08452 [cs, stat]},
}

@misc{becker_uncertainty_2022,
	title = {On {Uncertainty} in {Deep} {State} {Space} {Models} for {Model}-{Based} {Reinforcement} {Learning}},
	url = {http://arxiv.org/abs/2210.09256},
	doi = {10.48550/arXiv.2210.09256},
	urldate = {2026-09-03},
	publisher = {arXiv},
	author = {Becker, Philipp and Neumann, Gerhard},
	month = oct,
	year = {2022},
	note = {arXiv:2210.09256 [cs.LG]},
}

@article{silver_mastering_2017,
	title = {Mastering the game of {Go} without human knowledge},
	volume = {550},
	issn = {0028-0836, 1476-4687},
	url = {https://www.nature.com/articles/nature24270},
	doi = {10.1038/nature24270},
	language = {en},
	number = {7676},
	urldate = {2026-09-03},
	journal = {Nature},
	author = {Silver, David and Schrittwieser, Julian and Simonyan, Karen and Antonoglou, Ioannis and Huang, Aja and Guez, Arthur and Hubert, Thomas and Baker, Lucas and Lai, Matthew and Bolton, Adrian and Chen, Yutian and Lillicrap, Timothy and Hui, Fan and Sifre, Laurent and Van Den Driessche, George and Graepel, Thore and Hassabis, Demis},
	month = oct,
	year = {2017},
	pages = {354--359},
}

@article{Ha.etal_2020,
	title = {Learning to {Walk} in the {Real} {World} with {Minimal} {Human} {Effort}},
	volume = {abs/2002.08550},
	url = {https://arxiv.org/abs/2002.08550},
	journal = {CoRR},
	author = {Ha, Sehoon and Xu, Peng and Tan, Zhenyu and Levine, Sergey and Tan, Jie},
	year = {2020},
	note = {arXiv: 2002.08550},
}

@misc{luo_controlling_2022,
	title = {Controlling {Commercial} {Cooling} {Systems} {Using} {Reinforcement} {Learning}},
	copyright = {Creative Commons Attribution 4.0 International},
	url = {https://arxiv.org/abs/2211.07357},
	doi = {10.48550/ARXIV.2211.07357},
	urldate = {2026-09-03},
	publisher = {arXiv},
	author = {Luo, Jerry and Paduraru, Cosmin and Voicu, Octavian and Chervonyi, Yuri and Munns, Scott and Li, Jerry and Qian, Crystal and Dutta, Praneet and Davis, Jared Quincy and Wu, Ningjia and Yang, Xingwei and Chang, Chu-Ming and Li, Ted and Rose, Rob and Fan, Mingyan and Nakhost, Hootan and Liu, Tinglin and Kirkman, Brian and Altamura, Frank and Cline, Lee and Tonker, Patrick and Gouker, Joel and Uden, Dave and Bryan, Warren Buddy and Law, Jason and Fatiha, Deeni and Satra, Neil and Rothenberg, Juliet and Waraich, Mandeep and Carlin, Molly and Tallapaka, Satish and Witherspoon, Sims and Parish, David and Dolan, Peter and Zhao, Chenyu and Mankowitz, Daniel J.},
	year = {2022},
	note = {Version Number: 2},
}

@inproceedings{hafner_learning_2019,
	series = {Proceedings of {Machine} {Learning} {Research}},
	title = {Learning {Latent} {Dynamics} for {Planning} from {Pixels}},
	volume = {97},
	url = {https://proceedings.mlr.press/v97/hafner19a.html},
	booktitle = {Proceedings of the 36th {International} {Conference} on {Machine} {Learning}},
	publisher = {PMLR},
	author = {Hafner, Danijar and Lillicrap, Timothy and Fischer, Ian and Villegas, Ruben and Ha, David and Lee, Honglak and Davidson, James},
	editor = {Chaudhuri, Kamalika and Salakhutdinov, Ruslan},
	month = jun,
	year = {2019},
	pages = {2555--2565},
}

@article{moos_robust_2022,
	title = {Robust {Reinforcement} {Learning}: {A} {Review} of {Foundations} and {Recent} {Advances}},
	volume = {4},
	issn = {2504-4990},
	shorttitle = {Robust {Reinforcement} {Learning}},
	url = {https://www.mdpi.com/2504-4990/4/1/13},
	doi = {10.3390/make4010013},
	language = {en},
	number = {1},
	urldate = {2026-09-04},
	journal = {Machine Learning and Knowledge Extraction},
	author = {Moos, Janosch and Hansel, Kay and Abdulsamad, Hany and Stark, Svenja and Clever, Debora and Peters, Jan},
	month = mar,
	year = {2022},
	pages = {276--315},
}

@misc{pattanaik_robust_2017,
	title = {Robust {Deep} {Reinforcement} {Learning} with {Adversarial} {Attacks}},
	url = {http://arxiv.org/abs/1712.03632},
	doi = {10.48550/arXiv.1712.03632},
	urldate = {2026-03-19},
	publisher = {arXiv},
	author = {Pattanaik, Anay and Tang, Zhenyi and Liu, Shuijing and Bommannan, Gautham and Chowdhary, Girish},
	month = dec,
	year = {2017},
	note = {arXiv:1712.03632 [cs]},
}

@misc{huang_adversarial_2017,
	title = {Adversarial {Attacks} on {Neural} {Network} {Policies}},
	url = {http://arxiv.org/abs/1702.02284},
	doi = {10.48550/arXiv.1702.02284},
	urldate = {2026-09-04},
	publisher = {arXiv},
	author = {Huang, Sandy and Papernot, Nicolas and Goodfellow, Ian and Duan, Yan and Abbeel, Pieter},
	month = feb,
	year = {2017},
	note = {arXiv:1702.02284 [cs.LG]},
}

@inproceedings{todorov_mujoco_2012,
	title = {{MuJoCo}: {A} physics engine for model-based control},
	doi = {10.1109/IROS.2012.6386109},
	booktitle = {2012 {IEEE}/{RSJ} {International} {Conference} on {Intelligent} {Robots} and {Systems}},
	author = {Todorov, Emanuel and Erez, Tom and Tassa, Yuval},
	year = {2012},
	pages = {5026--5033},
}

@article{raffin_stable-baselines3_2021,
	title = {Stable-{Baselines3}: {Reliable} {Reinforcement} {Learning} {Implementations}},
	volume = {22},
	url = {http://jmlr.org/papers/v22/20-1364.html},
	number = {268},
	journal = {Journal of Machine Learning Research},
	author = {Raffin, Antonin and Hill, Ashley and Gleave, Adam and Kanervisto, Anssi and Ernestus, Maximilian and Dormann, Noah},
	year = {2021},
	pages = {1--8},
}

@misc{zhang_robust_2021-1,
	title = {Robust {Deep} {Reinforcement} {Learning} against {Adversarial} {Perturbations} on {State} {Observations}},
	url = {http://arxiv.org/abs/2003.08938},
	doi = {10.48550/arXiv.2003.08938},
	urldate = {2026-09-09},
	publisher = {arXiv},
	author = {Zhang, Huan and Chen, Hongge and Xiao, Chaowei and Li, Bo and Liu, Mingyan and Boning, Duane and Hsieh, Cho-Jui},
	month = jul,
	year = {2021},
	note = {arXiv:2003.08938 [cs.LG]},
}
\end{document}